\documentclass[a4paper,twoside]{article}

\usepackage{epsfig}
\usepackage{subcaption}
\usepackage{calc}
\usepackage{amssymb}
\usepackage{amstext}
\usepackage{amsmath}
\usepackage{amsthm}
\usepackage{multicol}
\usepackage{pslatex}
\usepackage{apalike}
\usepackage{algorithm2e}
\usepackage[bottom]{footmisc}

\usepackage{comment}
\usepackage{placeins}
\usepackage{flafter} 
\usepackage{float}
\usepackage{SCITEPRESS}     

\begin{document}

\title{Decision Making and Ride Comfort Evaluation Based on Lane-Change Prediction for Autonomous Vehicles}

\author{\authorname{Mohamed Manzour\sup{1}\orcidAuthor{0009-0007-2009-3573}, Aditya Kumar\sup{1}\orcidAuthor{0009-0004-5717-1027}, Rubén Izquierdo\sup{1}\orcidAuthor{0000-0002-6722-3036} and Miguel Ángel Sotelo\sup{1}\orcidAuthor{0000-0001-8809-2103 }}
\affiliation{\sup{1}Computer Engineering Department, University of Alcalá, Madrid, Spain}
\email{\{ahmed.manzour, aditya.kumar, ruben.izquierdo, miguel.sotelo\}@uah.es}
}


\abstract{Predicting surrounding vehicles' lane-change intentions enables autonomous vehicles to anticipate potential hazards and respond proactively rather than relying solely on reactive maneuvers. This study investigates the benefits of integrating lane-change prediction, based on knowledge graph embeddings and Bayesian inference, into autonomous vehicle decision-making. Three driving strategies are evaluated in risky merging scenarios simulated in CARLA: reactive driving, prediction-based longitudinal action, and prediction-based lateral action. The proposed prediction model anticipates the surrounding vehicle's lane change approximately 3-4 seconds before its occurrence, allowing the ego vehicle to adjust its behavior accordingly. In the representative simulation records, reactive driving results in a collision accompanied by aggressive deceleration. In contrast, prediction-based longitudinal action initiates early braking to establish a safe gap, while prediction-based lateral action performs a lane change when a safe adjacent lane is available. Both predictive strategies avoid collisions in the evaluated records. Compared with reactive driving, peak deceleration before impact is reduced by 31.5\% with longitudinal action and 91.0\% with lateral action. Additionally, an ISO 2631-1-based horizontal vibration estimate is used to provide a ride comfort assessment. Overall, the results highlight the potential of lane-change prediction to support safer and smoother driving through proactive decision-making, with additional insights into ride comfort provided by vibration-based estimates.}
\onecolumn \maketitle \normalsize \setcounter{footnote}{0} \vfill

\section{\uppercase{Introduction}}
\label{sec:introduction}
\begin{figure}[!htbp]
  \centering
  {\epsfig{file = 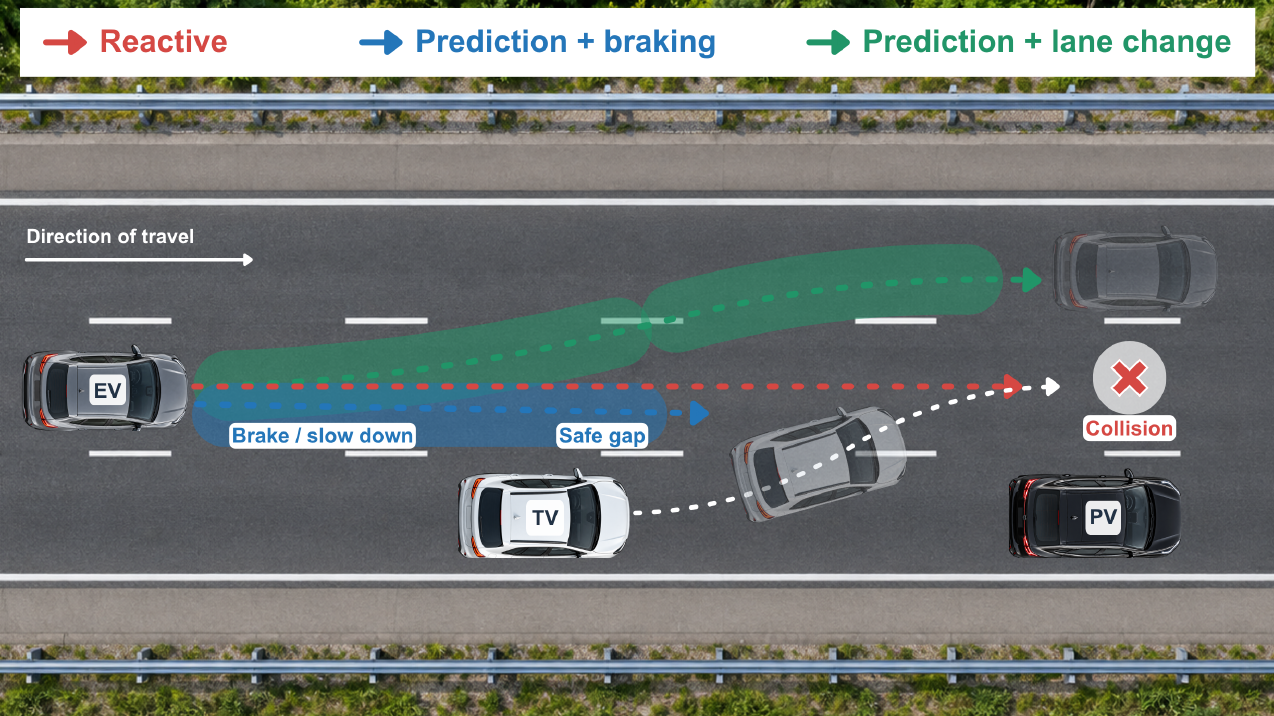, width = \linewidth}}
  \caption{The white target vehicle, following the black preceding vehicle, is predicted to merge into the grey ego vehicle's lane. Red: collision; blue: braking; green: lateral avoidance. Transparent vehicles indicate future positions.}
  \label{fig:intro}
\end{figure}

Lane changes are among the most common driving maneuvers, yet unexpected lane changes can create hazardous situations, particularly when a vehicle merges into an occupied lane with limited space for braking or avoidance. Predicting such maneuvers before they begin gives an autonomous vehicle additional time to respond. However, the benefit of this anticipation depends on the driving decision taken, as different responses may have different implications for safety and ride comfort. Figure~\ref{fig:intro} illustrates a scenario in which a target vehicle approaches a slower preceding vehicle in the rightmost lane and intends to merge into the ego vehicle's lane. Without anticipation, the ego vehicle may be unable to avoid a collision. With prediction, it can either brake to create a safe gap or change lanes when an adjacent lane is available.
While avoiding collisions is essential, the resulting vehicle motion also affects passenger comfort. Braking may introduce substantial longitudinal deceleration, whereas lateral avoidance produces acceleration in a different direction. Since discomfort depends on the magnitude, direction, and timing of vehicle motion \cite{Bellem2018,deWinkel2023}, these responses should be examined beyond their safety outcomes. This study investigates the effects of lane-change prediction on autonomous vehicle decision-making and ride comfort using an existing knowledge graph embedding and Bayesian inference framework \cite{ManzourIV2024,ManzourIV2025}. Three driving configurations are compared in CARLA: reactive driving, prediction-based longitudinal control, and prediction-based lateral control. Their behavior is analyzed through response timing, collision outcomes, acceleration, and jerk, while ride comfort is additionally assessed using an ISO 2631-1-based estimate of horizontal vibration.

\section{\uppercase{Related Work}}

Lane-change prediction has evolved from Long Short-Term Memory (LSTM)-based models capturing vehicle motion and traffic context \cite{Su2018} to dual-Transformer architectures combining intention and trajectory prediction \cite{Gao2023}. More recently, an interactive dynamic Bayesian network integrated with a Transformer was proposed to model vehicle interactions and jointly predict intentions and trajectories \cite{Hao2025}. Beyond prediction, planning based on Partially Observable Markov Decision Processes (POMDPs) has been investigated for decision-making under uncertain vehicle intentions \cite{Hubmann2017}. However, avoiding a collision does not necessarily ensure a comfortable driving response, particularly when comparing braking with lane changing. Previous research has investigated the influence of longitudinal and lateral acceleration and jerk on passenger discomfort \cite{deWinkel2023}.
 Complementing these measures, ISO 2631-1 provides a frequency-weighted vibration assessment framework \cite{ISO2631_1997}, also considered in simulation-based comfort evaluation \cite{Genser2022}. Nevertheless, the safety and comfort trade-offs between anticipatory braking and lateral avoidance in response to predicted lane changes remain insufficiently explored. To address this gap, this study compares both prediction-based strategies with reactive driving in CARLA, evaluating collision outcomes, response timing, vehicle dynamics, and ISO-based ride comfort.

\section{\uppercase{Methodology}}
\subsection{Lane-Change Prediction}
\label{subsec:lane_change_prediction}

\begin{figure}[!htbp]
  \centering
  {\epsfig{file = 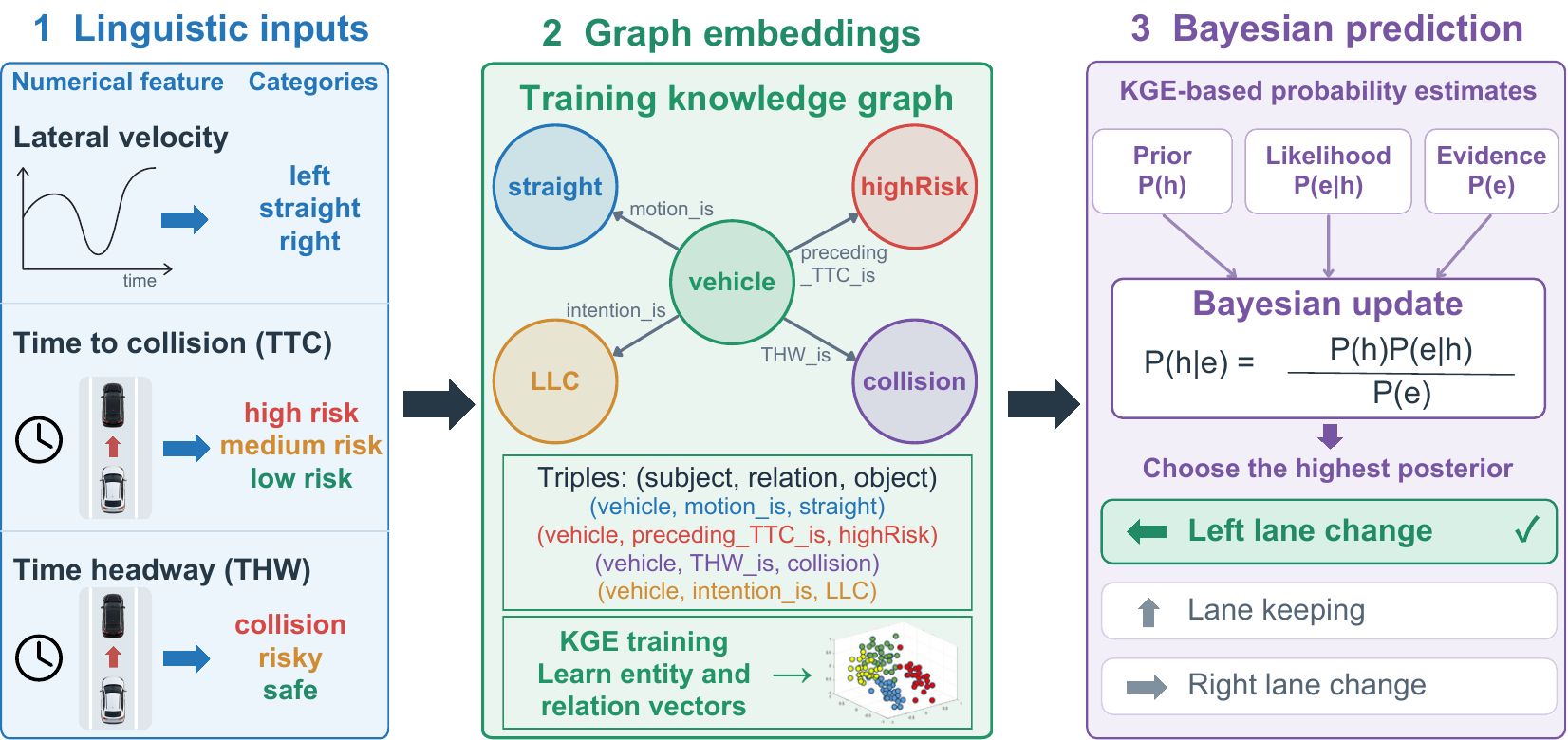, width = \linewidth}}
  \caption{Prediction stages: convert measurements into descriptive categories, learn graph embeddings, then use Bayesian inference to predict the maneuver. Illustrations are schematic.}
  \label{fig:method_framework}
\end{figure}

This study adopts the lane-change predictor presented in \cite{ManzourIV2025}, illustrated in Figure~\ref{fig:method_framework}. 
The model uses 12 features: lateral velocity and acceleration, lane identity, lane-marking offset, time headway, five Time-to-Collision (TTC) values with surrounding vehicles, and the lanes with the highest attraction score and largest frontal gap. Numerical inputs are converted into linguistic categories and represented as knowledge graph triples. TransE \cite{Bordes2013}, implemented in AmpliGraph \cite{AmpliGraph2019}, learns 100-dimensional embeddings from highD \cite{Krajewski2018} and CRASH datasets. Bayesian inference then uses these embeddings to predict left lane change, lane keeping, or right lane change. Further implementation details are provided in \cite{ManzourIV2025}.

\subsection{Decision Making}

The predicted lane-change is incorporated into the ego vehicle's decision-making process by assigning the target vehicle to its anticipated destination lane even before the target vehicle starts the maneuver. The risk associated with each lane is calculated as $R_{\ell}=\sum_{j\in\mathcal N_\ell}1/\mathrm{TTC}_{\ell j}$, where $R_{\ell}$ represents the lane risk score, $\mathcal N_\ell$ is the set of relevant surrounding vehicles in lane $\ell$, and $\mathrm{TTC}_{\ell j}$ is the Time-to-Collision with vehicle $j$. Only positive, finite closing TTC values are considered. The ego vehicle selects the lane with the lowest risk score, subject to lane availability. When lane keeping is selected, the Intelligent Driver Model (IDM) \cite{Treiber2000} determines the desired longitudinal acceleration based on vehicle speed, spacing, and relative velocity, while a PID controller generates throttle and braking commands. When a lane change is selected, Model Predictive Control (MPC) tracks a fifth-order lane-change trajectory, with IDM--PID governing longitudinal motion outside the maneuver. The reactive system also employs IDM--PID for longitudinal control. However, unlike the predictive system, it considers the target vehicle only after it enters the ego vehicle's lane, resulting in a later response.

\subsection{Experimental Setup}
\vspace{1mm}
The three driving configurations are evaluated in the CARLA simulator \cite{Dosovitskiy2017}. The simulated scenario involves a target vehicle approaching a slower preceding vehicle and executing a risky lane change toward the ego vehicle's lane. Reactive driving and anticipatory braking are evaluated on two-lane roads, while lateral avoidance uses a three-lane road to provide an available adjacent lane. Each configuration includes three runs, with the presented figures and numerical results obtained from a representative run. The evaluation considers collision outcomes, Time-to-Collision (TTC), braking response time, and longitudinal and lateral acceleration and jerk. Vehicle dynamics and ride comfort are compared over a common three-second interval preceding the reactive collision, while the complete simulation records illustrate the overall vehicle behavior.

\vspace{1mm}

\subsection{Ride Comfort Assessment}
\label{subsec:comfort_method}
\vspace{1mm}
Ride comfort is assessed using an ISO 2631-1-based estimate of horizontal vibration \cite{ISO2631_1997}. Longitudinal and lateral accelerations are frequency-weighted using the standard $W_d$ weighting, implemented through the phonometry library (version 3.3.0) \cite{RequenaPlens2026}, and combined as
\vspace{1mm}

\begin{equation}
a_{v,h}=\sqrt{\frac{1}{T}\int_{t_0}^{t_1}
\left([W_d\{a_x\}(t)]^2+[W_d\{a_y\}(t)]^2\right)\,dt},
\end{equation}
\vspace{1mm}

where $a_{v,h}$ is the combined weighted RMS acceleration, $a_x$ and $a_y$ represent longitudinal and lateral accelerations, respectively, $W_d$ is the frequency-dependent horizontal weighting function, and $T=t_1-t_0=3$ s is the common evaluation duration preceding the reactive collision. Further details on the weighting and calculation procedure are provided in \cite{ISO2631_1997,RequenaPlens2026}. Since acceleration is obtained from simulated vehicle motion rather than passenger-seat measurements, the assessment assumes that vehicle accelerations approximate those experienced by passengers. Vertical, rotational, and high-frequency vibrations are not fully represented, limiting the results to an indicative horizontal comfort estimate rather than a complete ISO 2631-1 assessment. The resulting values are interpreted using the approximate ISO 2631-1 discomfort ranges.

\section{\uppercase{Results and Discussion}}

\subsection{Safety and Response Timing}
\label{subsec:safety_results}

The prediction model anticipates the target vehicle's lane change approximately 2.5 s before crossing the lane marking in the selected scenario, while anticipation times of 3--4 s have been demonstrated in other scenarios \cite{ManzourIV2025}. In the reactive configuration, the ego vehicle collides with the merging target, whereas no collision occurs in either predictive configuration. Reactive braking begins only 0.97 s before the collision, compared with approximately 3.5 s for both predictive configurations, providing an additional 2.5 s of response time. The TTC results further illustrate these differences. Reactive driving reaches a minimum TTC of 0.015 s immediately before collision, compared with 3.9 s for prediction longitudinal. Prediction lateral reaches a minimum TTC of 1.3 s. However, since TTC considers only longitudinal closing motion, it does not account for the lateral separation between vehicles during lane changing. Therefore, a low TTC during lateral motion does not necessarily indicate an imminent collision.

\subsection{Vehicle Motion and Ride Comfort}
\label{subsec:comfort_results}

\begin{figure}[!htbp]
  \centering
  {\epsfig{file = 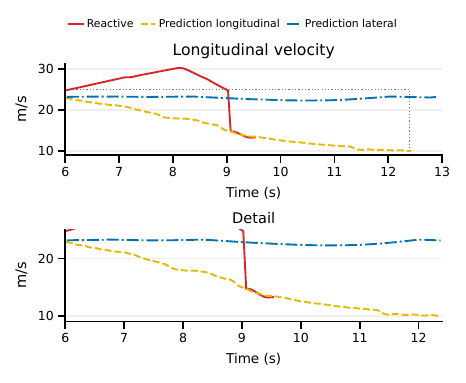, width = \linewidth}}
  \caption{Longitudinal velocity profiles. The reactive vehicle experiences a sudden speed reduction following impact, while anticipatory braking produces a gradual deceleration and lateral avoidance preserves forward motion. Upper: full range; lower: zoomed view.}
  \label{fig:comfort_vx}
\end{figure}

\begin{figure}[!htbp]
  \centering
  {\epsfig{file = 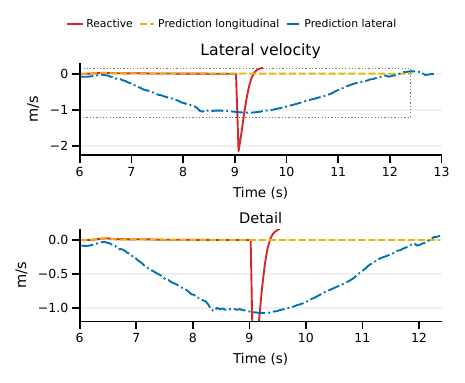, width = \linewidth}}
  \caption{Lateral velocity profiles. The avoidance maneuver (blue) introduces lateral movement toward the adjacent lane, whereas the reactive vehicle exhibits a lateral disturbance associated with the collision. Upper: full range; lower: zoomed view.}
  \label{fig:comfort_vy}
\end{figure}

\begin{figure}[!htbp]
  \centering
  {\epsfig{file = 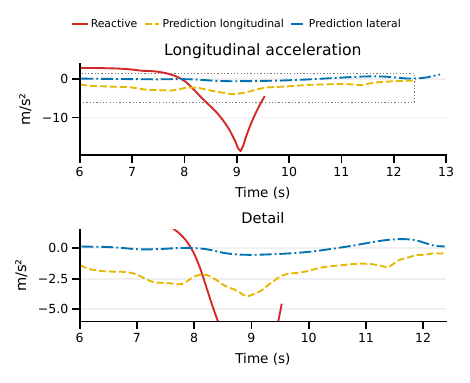, width = \linewidth}}
  \caption{Longitudinal acceleration profiles. The reactive response produces a sharp deceleration at impact, while early braking distributes deceleration over a longer interval. Lateral avoidance requires substantially less longitudinal braking. Upper: full range; lower: zoomed view.}
  \label{fig:comfort_ax}
\end{figure}

\begin{figure}[!htbp]
  \centering
  {\epsfig{file = 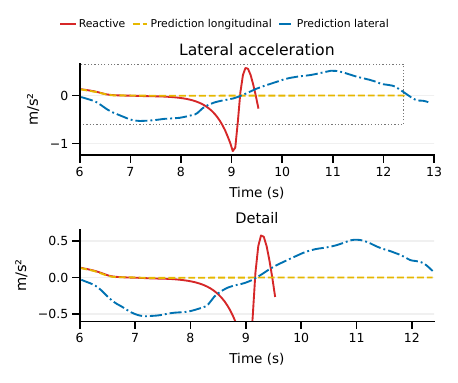, width = \linewidth}}
  \caption{Lateral acceleration profiles. Lane-change execution introduces lateral acceleration as the vehicle enters the adjacent lane. Upper: full range; lower: zoomed view.}
  \label{fig:comfort_ay}
\end{figure}

\begin{figure}[!htbp]
  \centering
  {\epsfig{file = 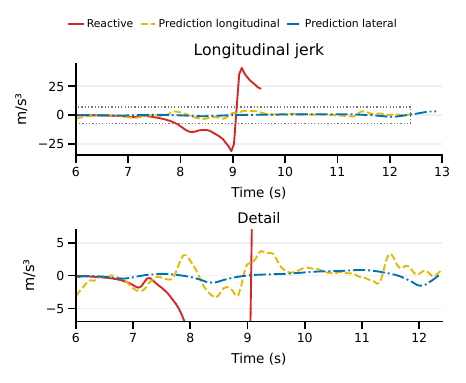, width = \linewidth}}
  \caption{Longitudinal jerk profiles. The largest variation occurs during the reactive collision, whereas anticipatory actions reduce abrupt changes in longitudinal acceleration. Upper: full range; lower: zoomed view.}
  \label{fig:comfort_jx}
\end{figure}

\begin{figure}[!htbp]
  \centering
  {\epsfig{file = 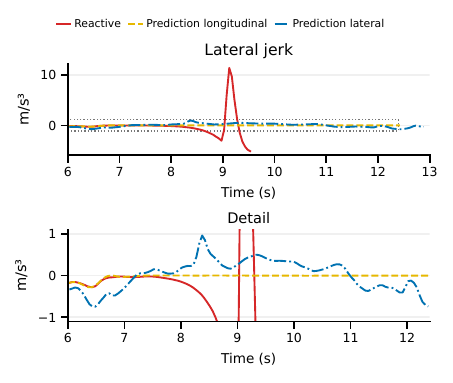, width = \linewidth}}
  \caption{Lateral jerk profiles. The avoidance maneuver introduces acceleration variations associated with lateral movement, while the reactive collision produces a clear spike. Upper: full range; lower: zoomed view.}
  \label{fig:comfort_jy}
\end{figure}

\begin{figure}[!htbp]
  \centering
  {\epsfig{file = 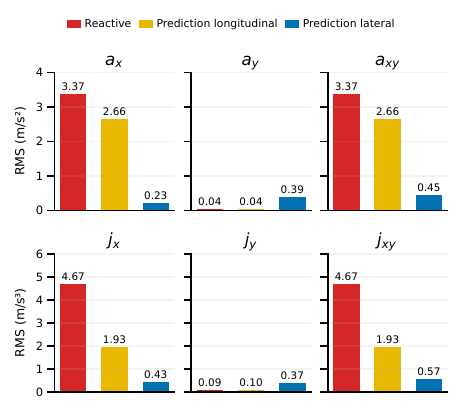, width = \linewidth}}
  \caption{Pre-impact RMS acceleration and jerk in the longitudinal, lateral, and combined directions. Although lane changing increases lateral motion, its substantially lower longitudinal demands result in the smallest combined RMS values. Each row shares the same vertical scale.}
  \label{fig:comfort_rms}
  
\end{figure}

Figures~\ref{fig:comfort_vx} and \ref{fig:comfort_vy} compare the longitudinal and lateral velocities under reactive driving, prediction-based longitudinal braking, and prediction-based lateral avoidance. The reactive vehicle experiences an abrupt speed reduction following the collision, whereas early braking allows a gradual deceleration. 
Alternatively, the avoidance lateral maneuver enables the ego vehicle to maintain a relatively stable longitudinal velocity while moving into the adjacent lane.
The corresponding acceleration and jerk profiles are presented in Figures~\ref{fig:comfort_ax}--\ref{fig:comfort_jy}. The collision produces a peak longitudinal deceleration of 18.8 m/s$^2$ and jerk of 41.0 m/s$^3$ in the reactive case. To distinguish the effects of the driving actions from the collision impact, Table~\ref{tab:comfort_metrics} compares the vehicle dynamics during the three seconds preceding the collision.
\begin{table}[!htbp]
\caption{Comparison of unweighted longitudinal and lateral acceleration and jerk during the three seconds preceding the reactive collision. \textit{Collision-induced motion is excluded from the table but included in the figures for visualization.}}
\label{tab:comfort_metrics}
\centering
\scriptsize
\setlength{\tabcolsep}{3pt}
\begin{tabular}{lrrr}
\hline
Metric & Reactive & \shortstack{Anticipatory\\braking} & \shortstack{Lateral\\avoidance}\\
\hline
Peak deceleration (m/s$^2$) & 6.427 & 4.406 & 0.578\\
Peak $|a_y|$ (m/s$^2$) & 0.131 & 0.130 & 0.530\\
Peak $|j_x|$ (m/s$^3$) & 11.853 & 4.116 & 1.087\\
Peak $|j_y|$ (m/s$^3$) & 0.275 & 0.279 & 0.882\\
$A_{x,\mathrm{RMS}}$ (m/s$^2$) & 3.366 & 2.658 & 0.230\\
$A_{y,\mathrm{RMS}}$ (m/s$^2$) & 0.037 & 0.037 & 0.388\\
$A_{xy,\mathrm{RMS}}$ (m/s$^2$) & 3.366 & 2.659 & 0.451\\
$J_{x,\mathrm{RMS}}$ (m/s$^3$) & 4.673 & 1.927 & 0.430\\
$J_{y,\mathrm{RMS}}$ (m/s$^3$) & 0.093 & 0.095 & 0.372\\
$J_{xy,\mathrm{RMS}}$ (m/s$^3$) & 4.674 & 1.929 & 0.569\\
\hline
\end{tabular}
\end{table}
Compared with reactive driving, anticipatory braking reduces pre-impact peak deceleration by 31.5\%, while lateral avoidance achieves a 91.0\% reduction. However, the latter introduces additional lateral motion, increasing RMS lateral acceleration from 0.037 to 0.388 m/s$^2$ and lateral jerk from 0.095 to 0.372 m/s$^3$ compared with prediction based braking. This trade-off highlights the importance of considering both motion directions when assessing maneuver smoothness \cite{deWinkel2023,Bellem2018}. 
Figure~\ref{fig:comfort_rms} compares the combined longitudinal and lateral RMS acceleration and jerk across the three strategies. Compared with reactive driving, combined RMS jerk decreases from 4.67 m/s$^3$ to 1.93 m/s$^3$ with anticipatory braking (58.7\% reduction) and to 0.57 m/s$^3$ with lateral avoidance (87.8\% reduction). Comparing the two prediction-based strategies directly, lateral avoidance reduces combined RMS acceleration from 2.66 to 0.451 m/s$^2$ (83.0\%) and combined RMS jerk from 1.93 to 0.57 m/s$^3$ (70.5\%). These findings demonstrate that lateral avoidance achieves the lowest overall acceleration and jerk levels, despite introducing additional lateral motion.

\subsection{ISO-Based Comfort}
\label{subsec:iso_results}

While Table~\ref{tab:comfort_metrics} evaluates vehicle motion using unweighted acceleration and jerk, Table~\ref{tab:iso} presents the RMS acceleration after applying the ISO 2631-1 horizontal frequency weighting ($W_d$), which accounts for human sensitivity to different vibration frequencies. The table also includes the approximate discomfort ranges specified in Annex C of the ISO 2631-1 standard \cite{ISO2631_1997}.
\begin{table}[!htbp]
\vspace{2mm}
\caption{ISO 2631-1 horizontal frequency-weighted RMS acceleration during the three seconds preceding the reactive collision, alongside the approximate discomfort ranges from Annex C of the standard \cite{ISO2631_1997}. All acceleration values are expressed in m/s$^2$.}
\label{tab:iso}
\centering
\scriptsize
\setlength{\tabcolsep}{3pt}
\begin{tabular}{lrrr}
\hline
Driving strategy & $a_{wx}$ & $a_{wy}$ & $a_{v,h}$\\
\hline
Reactive driving & 1.51 & 0.17 & 1.52\\
Anticipatory braking & 1.41 & 0.16 & 1.42\\
Lateral avoidance & 0.48 & 0.23 & 0.53\\
\hline
\multicolumn{3}{l}{ISO 2631-1 Annex C discomfort ranges} & RMS range\\
\hline
\multicolumn{3}{l}{Not uncomfortable} & $<0.315$\\
\multicolumn{3}{l}{A little uncomfortable} & 0.315--0.63\\
\multicolumn{3}{l}{Fairly uncomfortable} & 0.5--1.0\\
\multicolumn{3}{l}{Uncomfortable} & 0.8--1.6\\
\multicolumn{3}{l}{Very uncomfortable} & 1.25--2.5\\
\multicolumn{3}{l}{Extremely uncomfortable} & $>2.0$\\
\hline
\end{tabular}
\end{table}
Reactive driving exhibits the highest pre-impact combined frequency-weighted RMS acceleration of 1.52 m/s$^2$, which decreases slightly to 1.42 m/s$^2$ with anticipatory braking. In contrast, lateral avoidance achieves 0.53 m/s$^2$, representing reductions of 65\% and 63\% compared with reactive driving and anticipatory braking, respectively. According to the approximate discomfort ranges provided in ISO 2631-1 Annex C, the first two values lie within the overlapping ``uncomfortable'' and ``very uncomfortable'' ranges, whereas lateral avoidance falls within the overlapping ``a little uncomfortable'' and ``fairly uncomfortable'' ranges. These results suggest that lateral avoidance provides the most favorable frequency-weighted acceleration characteristics in the evaluated scenario. However, the assessment considers only horizontal vehicle accelerations rather than complete passenger-seat vibration measurements. The ISO discomfort ranges are therefore indicative rather than definitive.

\section{\uppercase{Conclusion}}

This study evaluated the impact of lane-change intention prediction on driving safety and ride comfort through longitudinal braking and lateral avoidance actions. Prediction enabled the ego vehicle to respond approximately 2.5 s earlier than reactive driving, preventing the collision observed in the reactive scenario. During the pre-impact comparison interval, anticipatory braking and lateral avoidance reduced peak longitudinal deceleration by 31.5\% and 91.0\%, respectively, and combined RMS jerk by 58.7\% and 87.8\%. Compared with braking alone, lateral avoidance further reduced peak deceleration by 86.9\% and combined RMS jerk by 70.5\%, despite introducing additional lateral motion. The ISO-based assessment further supported these findings, with lateral avoidance exhibiting the lowest frequency-weighted acceleration and the most favorable estimated comfort level. Overall, the results demonstrate the potential of integrating lane-change prediction into driving decisions, allowing the ego vehicle to select between early braking and lateral avoidance according to the surrounding traffic conditions and available space. Nevertheless, the findings are based on simulated scenarios and vehicle-level acceleration measurements. Further experiments across diverse traffic conditions, together with passenger-seat vibration measurements, are required to validate the generalizability of the results and their implications for perceived ride comfort.

\bibliographystyle{apalike}
{\small\bibliography{example}}
\end{document}